%% file: main.tex
\documentclass{article}

\PassOptionsToPackage{numbers,sort&compress}{natbib}
\usepackage[dblblindworkshop,final]{neurips_2026}
\workshoptitle{On-Device Intelligence: Foundation Models under Real-World Constraints}
\makeatletter
\renewcommand{\@trackname}{NeurIPS \@neuripsyear\ Workshop on \@workshoptitle.}
\makeatother

\usepackage[utf8]{inputenc}
\usepackage[T1]{fontenc}
\usepackage{amsmath,amsfonts,amssymb}
\usepackage{booktabs}
\usepackage{graphicx}
\usepackage{microtype}
\usepackage{xcolor}
\usepackage{array}
\usepackage{tabularx}
\usepackage{url}
\usepackage{float}
\usepackage{placeins}
\usepackage{hyperref}

\definecolor{placeholder}{RGB}{163,38,56}
\definecolor{devicedraftblue}{RGB}{0,82,155}
\newcommand{\pp}{\,\mathrm{pp}}
\newif\ifshowplaceholderids
\ifdefined\MacTeamPreview
  \showplaceholderidstrue
\else
  \showplaceholderidsfalse
\fi
\newcommand{\result}[1]{%
  \ifshowplaceholderids
    \textcolor{placeholder}{\nolinkurl{[#1]}}%
  \else
    \textcolor{placeholder}{---}%
  \fi}
\newif\ifdeviceresults
\deviceresultsfalse
\def\AcceptedDeviceResults{}
\ifdefined\AcceptedDeviceResults
  \deviceresultstrue
\fi
\ifdefined\MacTeamPreview
  \deviceresultstrue
\fi
\newif\ifdevicedraftpreview
\devicedraftpreviewtrue
\newenvironment{DeviceDraftBlock}{%
  \begingroup
  \ifdefined\AcceptedDeviceResults
  \else
    \ifdevicedraftpreview
      \color{devicedraftblue}%
    \else
      \ifdefined\MacTeamPreview
        \color{devicedraftblue}%
      \fi
    \fi
  \fi
}{\endgroup}
\newcommand{\DeviceDraftMarker}{%
  \ifdefined\AcceptedDeviceResults
  \else
    \ifdevicedraftpreview
      \textbf{Device draft pending validation. }%
    \else
      \ifdefined\MacTeamPreview
        \textbf{Device draft pending validation. }%
      \fi
    \fi
  \fi
}
\newcommand{\DeviceDraftColor}{%
  \ifdefined\AcceptedDeviceResults
  \else
    \ifdevicedraftpreview
      \color{devicedraftblue}%
    \else
      \ifdefined\MacTeamPreview
        \color{devicedraftblue}%
      \fi
    \fi
  \fi
}

\def\PaperTitle{Scaling Context Down for On-Device Tabular Foundation Models}
\title{\PaperTitle}
\newcommand{\corr}{\textsuperscript{\normalfont\dag}}
\author{%
  Shuting Luo\corr \quad Henry Xiao\corr \quad Luiz Pizzato \quad Ruth Oliver \\
  Commonwealth Bank of Australia \\
}
\providecommand{\FigureOneCaption}{%
\textbf{Predictive performance at C128.}
C128 denotes exactly 128 labelled examples. Datasets are ordered by three-model mean C128 $\Delta$AUC and
aligned across panels. (A) how much performance remains: paired full-context and C128 ROC AUC, averaged over the three TFMs.
(B) how sensitive each task is: model-specific $\Delta$AUC (full-context AUC minus C128 AUC, in pp),
which also shows that the three TFMs rank the sensitive tasks similarly. The 5-pp line is illustrative, and a small
$\Delta$AUC does not imply high absolute AUC or an easy task.}

\begin{document}
\maketitle

{\renewcommand{\thefootnote}{}%
\footnotetext[0]{$\dag$~Corresponding authors: \texttt{aria.luo@cba.com.au},
\texttt{henry.xiao@cba.com.au}.}}

\begin{abstract}
Tabular foundation models (TFMs) make predictions through in-context learning by conditioning on labelled examples supplied at inference time. Processing this context can strain compute and memory on constrained devices. We ask whether a TFM needs its full context, or whether a small fraction can retain most predictive performance.
We evaluate three pretrained TFMs (TabPFN v2.6, TabPFN v3, and TabICL v2) on 32 TabArena classification tasks. Reducing the context to 128 examples keeps a majority of model--task evaluations within five AUC percentage points of full context, using less than 2\% of the full context for the median task. The three models agree strongly on which tasks are most sensitive.
We profile these 128-example runs on a MacBook Air M2 (24GB). All complete, while 12 full-context runs hit memory or time limits. For TabICL v2, among completed comparisons, the reduced context runs faster, in only 14.4\% of the full-context time and 32.1\% of its peak memory (medians). These results reframe deployment from how much context a model can accept to how much a task needs, while motivating the next question of which examples are worth retaining.

\end{abstract}

\section{Introduction}

Tabular foundation models predict from labelled examples supplied as task context
\citep{hollmann2025tabpfn,qu2025tabicl,qu2026tabiclv2,grinsztajn2026tabpfn3}. Unlike a conventional
fitted predictor, a TFM keeps conditioning on these task-specific labelled examples every time it
makes a prediction. Some implementations process the examples directly and others cache intermediate
representations; regardless, the context stays part of the inference procedure.

Carrying these labelled examples through every prediction poses a resource constraint for
on-device intelligence. A larger context means more
task-specific data to store and process at inference, which raises memory use and inference cost
\citep{zabergja2026taco,jamalidinan2026arash}. One way to lower that cost is to compress the model
itself, through quantisation \citep{luo2026memory} or by distilling a TFM into a smaller
dataset-specific model \citep{grinsztajn2026tabpfn25}. The inference-time context is a separate
target for the same goal, and it is the one we study.

A model may be able to consume, say, 10{,}000 labelled examples, but that does not mean every
task needs all 10{,}000. We are interested in the context efficiency of a task, how much predictive
performance remains when its context is smaller, and in how much this performance varies across tasks. We refer to a task as \emph{sensitive} when
shrinking its context causes a larger drop in AUC relative to full context. Our central question follows: \emph{how does
predictive performance respond as the labelled inference context is reduced, and do different TFMs agree
on which tasks lose most when they are given identical context?} We study TabArena
\citep{erickson2025tabarena}, where each task is a distinct, curated dataset.

Prior work already examines context efficiency and its quality--cost trade-off through compression,
retrieval, optimisation, long-context execution, context-size studies, and prototype or
query-specific selection
\citep{feuer2023scaling,ma2024icd,feuer2024tunetables,thomas2024localpfn,xu2024mixturepfn,
chen2026vipcop,zabergja2026taco,abdullah2026contextsampling,ipas2026lucos,jadid2026baps,
jamalidinan2026arash,lin2026tlandi,lee2026cure}. More directly, TabDPT varies retrieved context size
and reports predictive performance and runtime \citep{ma2025tabdpt}; TabFlex varies retained-row
fractions and reports predictive \mbox{performance} and latency \citep{zeng2025tabflex}; and Chunked TabPFN
reports per-dataset predictive-quality, runtime and memory curves over context length, including
task-dependent responses \citep{sergazinov2025chunked}. Recent work on Localized TabICLv2 likewise
varies query-specific retrieval size and measures accuracy retention and prediction time
\citep{guta2026localizedtabiclv2}. We therefore do not claim context reduction or the general
performance--cost trade-off as novel. Our narrower contribution is a controlled cross-model
characterisation: for each task, context size, and repeat, all three TFMs receive the same selected
rows, allowing us to compare task sensitivity and profile reuse under identical evidence; we then
profile C128 against full context on a consumer device.

\paragraph{Contributions.}
First, across 32 TabArena tasks and three pretrained
TFMs, we measure how predictive performance changes over context sizes of 16--512 examples
(written as \emph{C16--C512}) selected from the original labelled examples by a fixed model-agnostic
policy, relative to full context. We find substantial
variation across tasks but strong cross-model agreement in context sensitivity.
Second, we profile a context size of 128 (C128) against full context on a MacBook Air M2 (24GB), reporting completion, fit + predict
time and peak RSS.

\section{Experimental setup}

\paragraph{Data and shared contexts.}
We study the 32 TabArena classification tasks \citep{erickson2025tabarena} large enough to supply
every context size up to C512, excluding the 6 smaller classification datasets for this size reason, not on observed performance. Each experiment runs the
same pipeline. For a task $d$, we (1) use TabArena's provided train/test split;
(2) select exactly $C$ labelled examples from the train split using a model-agnostic policy;
(3) give that same selected context to pretrained TabPFN~v3, TabPFN~v2.6
and TabICL~v2; (4) evaluate every model on the same held-out test partition; and
(5) compare each model's result with its own full-context reference. Supplying context never updates model
parameters; we do no retraining, fine-tuning or gradient-based adaptation.
This provided split is fixed. For each task and context size, all three models receive the same
selected context, so any performance gap reflects the models rather than differing inputs. Because
the selection procedure is not fully deterministic, we repeat the selection and report each model's
AUC averaged over repeats.

\paragraph{Selection policy.}
For a task $d$ and context size $C$, a model-agnostic policy returns exactly $C$ original examples from
the train split. Our primary policy, \emph{balanced-energy selection}, first sets balanced per-class
quotas (adjusted for class availability), then fills each class's quota with a
greedy rule. This rule starts from the class example closest to the class mean and repeatedly adds the
one remaining example that most reduces the empirical statistical energy distance to that class
distribution in the train split, without revising earlier picks. This energy distance is defined in
Eq.~\ref{eq:app-energy} (Appendix~\ref{sec:app-selector}) and follows energy-distance representative
designs \citep{szekely2013energy,mak2018supportpoints}.

\paragraph{Absolute performance and \texorpdfstring{$\Delta$AUC}{Delta AUC}.}
We report AUC values directly and also measure the cost of reducing context as the drop in AUC
relative to full context, which we call \emph{$\Delta$AUC}. This is calculated as the full-context AUC minus the
reduced-context AUC, reported in percentage points (pp), so that positive $\Delta$AUC means the
reduced context scores lower. For a task $d$ and model $m$ (one of the three evaluated TFMs), with
$\mathrm{AUC}_{d,m}(C)$ the repeat-averaged score at context size $C$ is given by:
\begin{equation}
\Delta\mathrm{AUC}_{d,m}(C)=100\bigl[\mathrm{AUC}^{full}_{d,m}-\mathrm{AUC}_{d,m}(C)\bigr]\pp.
\label{eq:loss}
\end{equation}
A $\Delta$AUC of $3\pp$ means the reduced context lowers AUC by three percentage points - positive
values mean reduction of performance; negative values mean the reduced-context AUC exceeded
the full-context reference. This is possible because full context is a reference point, not
a guaranteed best case. We use one fixed reference AUC per task and model
across all context sizes, so every context size is compared with the same baseline. We report binary
ROC AUC, and macro-averaged one-vs-rest ROC AUC for multiclass tasks, over the exact grid
$\{16,32,64,128,256,512\}$. Full context supplies the complete train split
(902--100,000 examples) and each context size $C$ means exactly $C$ \emph{distinct original} examples.

\paragraph{Statistical analysis.}
To measure cross-model agreement, we rank tasks by $\Delta$AUC and correlate two models' rankings
with Spearman $\rho$ at each context size, over the tasks available there. We report 95\% confidence
intervals from bootstrap resampling of the task set. As a robustness check, we use a leave-one-model-out analysis: we hold out one TFM, average the other
two models' $\Delta$AUC profiles into a single profile, and correlate the held-out model's task ranking
against it, repeating for each held-out model. The full protocol is in Appendix~\ref{sec:app-agreement}.

\begin{DeviceDraftBlock}
\paragraph{On-device profiling setup.}
\label{sec:device-setup}
We also profile C128 against the full-context reference on a
MacBook Air with M2 and 24\,GB unified memory, for each of the three models, reusing the
same fixed model-agnostic C128 context per task. We run TabPFN~v2.6 and TabPFN~v3 using the MPS (Metal Performance Shaders) backend, and TabICL~v2 using CPU
after MPS execution failed with a consistent out-of-memory error when loading--full protocol in Appendix~\ref{sec:app-device}. We summarise cost results with the
\emph{resource ratio} (RR), the C128 cost divided by the full-context cost on the same task and
expressed as a percentage, so lower is cheaper: an RR of $10\%$ means C128 uses one tenth of the
full-context resource. Each timed run repeats setup and prediction three times and reports the
median. However, if a single pass exceeds 35 minutes we keep only that one pass (extra repeats contribute little once runtime dominates timing variance) and runs are capped at 120 minutes per task (conservative limit for on-device experimentation) given hardware limitations.
\end{DeviceDraftBlock}

\section{Results}

\begin{figure}[H]
  \centering
  \includegraphics[width=\linewidth]{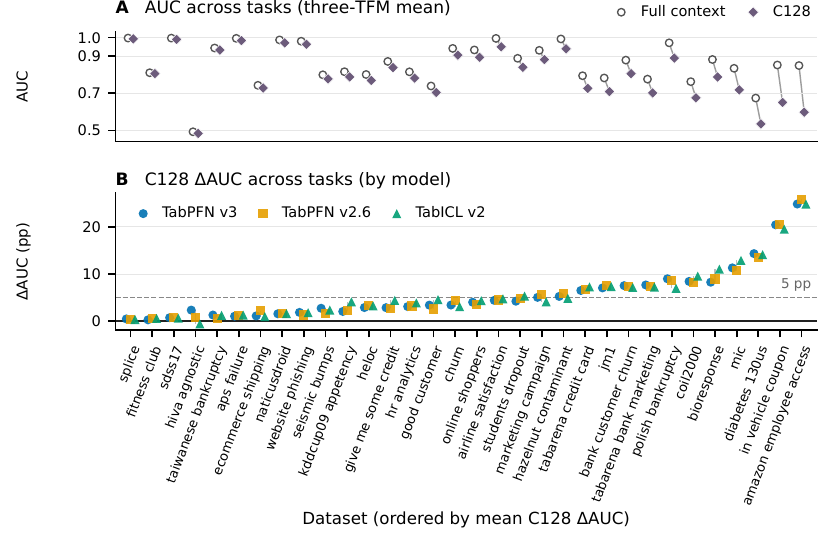}
  \caption{\FigureOneCaption}
  \label{fig:snapshot}
\end{figure}

\paragraph{Performance at C128.}
At C128, most tasks lose little predictive performance. Figure~\ref{fig:snapshot} compares C128 with the
full-context reference across the 32 tasks. We use C128 as a compact reference point - it is small enough
to ease on-device inference while keeping most of the full-context performance. Because the train sets contain 902--100,000 examples,
moving to C128 cuts the number of supplied rows to $0.13\%$--$14.2\%$ (median $1.9\%$) of the full context size. Even so, 58 of
96 task and model combinations stay within $5\pp$ of their full-context reference (the $5\pp$ line
is illustrative). Averaged over all evaluations, mean AUC is 0.860 with full context and 0.803 at
C128. The effect is far from uniform: individual task and model mean $\Delta$AUC ranges from $-0.57$ to
$25.79\pp$. Yet the three TFMs agree strongly on which tasks lose most: pairwise task-rank
correlations are $0.949$--$0.978$, with all three 95\% confidence intervals above zero
(Appendix Figure~\ref{fig:app-budget-agreement} and Table~\ref{tab:app-budget-pairwise}).

\FloatBarrier

\begin{figure}[!ht]
  \centering
  \includegraphics[width=\linewidth]{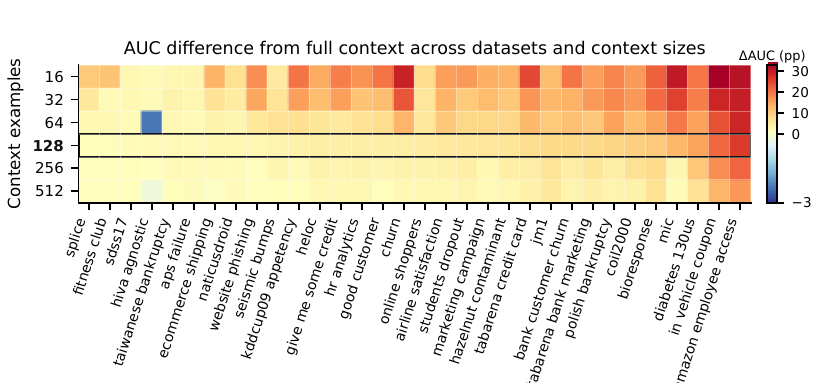}
  \caption{\textbf{$\Delta$AUC across tasks and context sizes.}
  The context-size-by-task heatmap uses exactly the Figure~\ref{fig:snapshot} task order, so every
  task can be followed from C16 through C512; the outlined C128 row links the two figures. Each
  cell is the equal-weight three-model mean $\Delta$AUC. Warm colours mark lower AUC under reduced
  context. Blue marks a negative $\Delta$AUC, where the reduced-context mean AUC exceeds the fixed
  full-context reference. A negative cell is not evidence that less context reliably helps; the
  prominent Hiva C64 cell is a near-chance, high-variance case examined in
  Appendix~\ref{app:hiva-negative-loss}. Cross-model rank-agreement estimates and their uncertainty
  appear in Appendix Figure~\ref{fig:app-budget-agreement} and
  Table~\ref{tab:app-budget-pairwise}.}
  \label{fig:profiles}
\end{figure}

\paragraph{Performance across context sizes.}
Figure~\ref{fig:profiles} shows how $\Delta$AUC changes from C16 to C512 for the 32 tasks, averaged over the three models.
On average, $\Delta$AUC shrinks with more context, but the gain is very uneven across tasks. All 32
tasks have enough train rows for every context size in the main grid, so the comparison is matched throughout.
Average $\Delta$AUC falls from $14.72\pp$ at C16 to $2.36\pp$ at C512, yet even at C512 the task means
still span $-0.30$ to $15.30\pp$ (the one visible negative case, Hiva at C64, is a high-variance
result possibly linked to its severe class imbalance, examined in Appendix~\ref{app:hiva-negative-loss}). The cross-model
agreement holds at every context size: across the 18 model-pair and context-size comparisons,
$\rho=0.912$--$0.983$ and every 95\% bootstrap interval excludes zero (smallest lower endpoint
0.753; Table~\ref{tab:app-budget-pairwise}). The leave-one-model-out analysis, where each model is compared against the mean profile of the other two, agrees, with median correlations across the context sizes of $0.942$--$0.965$ (Appendix~\ref{sec:app-agreement}).

\paragraph{Adjustment for full-context performance.}
One possibility is that the TFMs only agree because easier tasks, where full-context AUC is
already high, have less room to lose when context shrinks. To check this, we remove each model's
own full-context AUC as a predictor of its $\Delta$AUC, then measure agreement on what remains.
This adjusted agreement stays high, both at C128 ($\rho=0.944$--$0.975$) and across the full grid
($\rho=0.909$--$0.982$), so full-context performance alone does not explain the cross-model
ordering. We show this pattern holds after this check, not what causes it.
\paragraph{Classical ML controls.}
If the ordering were a generic property of the selected subsets, classical models trained
on the same examples should rank the tasks similarly. Summarising each task by mean $\Delta$AUC across the six context sizes, the three TFMs correlate strongly with one another ($\rho=0.957$--$0.970$), but logistic-regression controls correlate with the TFMs much more weakly ($0.293$--$0.378$) and shallow-tree controls show essentially no correlation ($-0.058$ to $-0.004$). This does not prove a uniquely TFM mechanism, but it makes the shared ordering hard to attribute to the selected rows alone.
\paragraph{Effect of context-selection policy.}
Changing the selector shifts the $\Delta$AUC values, but balanced allocation still beats
proportional, and random selection within each class ranks tasks nearly the same way as the
energy-distance heuristic. At C128, balanced class allocation gives lower mean $\Delta$AUC than
proportional allocation, both for energy-based selection ($5.67\pp$ vs.\ $8.02\pp$) and for random
selection ($6.03\pp$ vs.\ $8.49\pp$); within balanced allocation, energy-based and random selection
still induce similar TabPFN~v2.6 task orderings at C64, C128, and C512 ($\rho=0.942$--$0.979$). How
classes are allocated thus matters more than the within-class rule (full factorial and
multi-context-size controls in Appendix~\ref{sec:app-selector}).

\begin{DeviceDraftBlock}
\paragraph{On-device performance.}
\DeviceDraftMarker
We profile C128 on a MacBook (setup above). Table~\ref{tab:device} reports RR
for runtime and peak RSS. Every
completed task--model comparison (run under both C128 and full context) was faster and used less peak
memory at C128 than at full context. Median runtime RR was $6.6\%$, $8.9\%$, and $14.4\%$. Memory fell much less for the two TabPFN models (median RSS RR
$93.6\%$ and $95.0\%$) than for TabICL~v2 ($32.1\%$). C128 also enabled inference on all
tasks that full context failed to complete due to out-of-memory or timeout: 3 for TabPFN~v2.6, 1 for TabPFN~v3, and 8 for TabICL~v2,
excluded from the RR calculation; Appendix~\ref{sec:app-device} lists which tasks failed and for which reason. Furthermore, as TabICL v2 ran on CPU, peak process RSS may have captured context-related memory differently than for the two TabPFN models running via MPS. However, a separate imitative experiment on an A100 GPU confirms a similar RR value (Appendix~\ref{sec:app-device}).
\begin{table}[!ht]
\DeviceDraftColor
\caption{\textbf{MacBook C128 resource ratios (RR).} RR is the C128 cost as a percentage of the
full-context cost on the same task, so lower is cheaper. For each model we report the median,
P25, and P95 of RR over completed comparisons. Runtime combines context fit and query predict times
and excludes model loading; RSS is peak process resident memory. Each statistic is a
median-of-ratios-of-medians: first, three repeated runs are calculated as a per-task median; second RR is calculated per task; finally, the reported median, P25, and P95 are taken over
these ratios across the $n$ tasks.}
\label{tab:device}
\centering
\small
\setlength{\tabcolsep}{6pt}
\renewcommand{\arraystretch}{1.08}
\begin{tabular}{@{}lccc@{}}
\toprule
Model & $n$ tasks & \begin{tabular}[c]{@{}c@{}}Runtime RR \%\\{}median [P25, P95]\end{tabular}
& \begin{tabular}[c]{@{}c@{}}RSS RR \%\\{}median [P25, P95]\end{tabular}\\
\midrule
TabPFN v2.6 & 29 & 6.6 [2.8, 42.4]  & 93.6 [81.0, 99.2]\\
TabPFN v3   & 31 & 8.9 [2.2, 52.1]  & 95.0 [91.1, 99.2]\\
TabICL v2   & 24 & 14.4 [4.5, 36.0] & 32.1 [24.0, 67.0]\\
\bottomrule
\end{tabular}
\\[1pt]\parbox{0.98\linewidth}{\footnotesize The full-context reference did not complete for 3
TabPFN~v2.6 tasks, 1 TabPFN~v3 task, and 8 TabICL~v2 tasks; ratio calculations exclude those
tasks. Appendix~\ref{sec:app-device} lists these datasets and their failure type (out-of-memory or
timeout).}
\end{table}
\end{DeviceDraftBlock}

\section{Conclusion, Limitations, and Future Work}
We asked whether TFMs can use only a fraction of their inference context rows while preserving predictive
performance. Across 32 TabArena classification tasks and three pretrained TFMs (TabPFN v2.6, TabPFN
v3, and TabICL v2), reducing context to 128 examples (a median of under 2\% of the full context across tasks) keeps most
evaluations within five AUC points of full context. Context need varies sharply by task, yet the
three TFMs strongly agree on which tasks are most sensitive, with rank correlations above 0.91 at
every context size. This suggests context can be right-sized per task using profiles that are
largely consistent across models.

Since our contribution measures this phenomenon rather than proposing a new selection technique,
limitations arise as we intentionally used simple selection policies, giving a
conservative floor and a discrete context grid that bounds rather than identifies each task's
minimum. Additionally, we study only classification and three TFMs, two of which from the TabPFN family. Device results are specific to one MacBook Air M2 (24GB) and matched completed runs, where TabICL v2 ran on CPU after MPS
execution failed with an out-of-memory error at model load. Additionally, the MacBook Air M2 is a 2022 consumer device, the only one available to us for this paper.

Future work should optimise context selection, estimate task compressibility from cheap probes, and
develop adaptive right-sizing policies \citep{jamalidinan2026arash,lee2026cure}. Furthermore, utilising more TFMs drawn from distinct model families and benchmarking with tasks
from other open benchmarks such as OpenML \citep{vanschoren2014openml} would show how far these
patterns reach beyond the models and tasks in this paper. Additionally, benchmarking on
other devices would reveal how well these results generalise across hardware, such as more recent devices with more memory that can host larger models.

In practice, 128 examples provide a strong starting point, with context increased for sensitive tasks
and profiles reusable across models. Testing on the MacBook, C128 completed every task, including the 12 the full context failed due to
time and memory limits, and reduced time and memory in every completed comparison. Thus, on the
constrained device we tested, context right-sizing turned otherwise impractical inference into practical execution.

\label{page:mainend}

\clearpage
\small
\bibliographystyle{unsrtnat}
\bibliography{references}

\clearpage
\normalsize
\appendix
\input{appendix.tex}

\end{document}

%% file: appendix.tex
\section{Datasets, evaluation grid, and reproducibility}

\paragraph{Tasks and context sizes.}
The study uses 32 fixed TabArena classification tasks \citep{erickson2025tabarena}, listed in
Table~\ref{tab:app-task-list}.
C128 denotes a context of exactly 128 labelled examples, and the other C-prefixed sizes follow the
same convention. Training pools range from 902 to 100,000 rows, 6 to 1,776 features, and 2 to 8
classes, and every task supports all main-grid context sizes $C\in\{16,32,64,128,256,512\}$, each
drawn as exactly $C$ distinct rows. The three TFMs are each evaluated with three seeded
context-selection repeats under our primary policy (balanced allocation with greedy statistical-energy
selection), giving 1,728 evaluations and 576 aggregates over task, model, and context size.

\paragraph{Inference protocol.}
We evaluate all TFM checkpoints without task-specific parameter optimisation: supplying the context
only prepares it for in-context prediction and never retrains or fine-tunes the pretrained weights.
The logistic-regression and decision-tree controls in Table~\ref{tab:app-classical-control} are
conventional fitted learners, treated separately.

\begin{table}[h]
\caption{Fixed primary task population. All 32 tasks are retained at every main-grid context size; no task
is removed on the basis of observed performance.}
\label{tab:app-task-list}
\centering
\scriptsize
\begin{tabular}{ll}
\toprule
Task ID & Task ID\\
\midrule
\texttt{airline\_satisfaction} & \texttt{in\_vehicle\_coupon}\\
\texttt{amazon\_employee\_access} & \texttt{jm1}\\
\texttt{aps\_failure} & \texttt{kddcup09\_appetency}\\
\texttt{bank\_customer\_churn} & \texttt{marketing\_campaign}\\
\texttt{bioresponse} & \texttt{mic}\\
\texttt{churn} & \texttt{naticusdroid}\\
\texttt{coil2000} & \texttt{online\_shoppers}\\
\texttt{diabetes\_130us} & \texttt{polish\_bankruptcy}\\
\texttt{ecommerce\_shipping} & \texttt{sdss17}\\
\texttt{fitness\_club} & \texttt{seismic\_bumps}\\
\texttt{give\_me\_some\_credit} & \texttt{splice}\\
\texttt{good\_customer} & \texttt{students\_dropout}\\
\texttt{hazelnut\_contaminant} & \texttt{tabarena\_bank\_marketing}\\
\texttt{heloc} & \texttt{tabarena\_credit\_card}\\
\texttt{hiva\_agnostic} & \texttt{taiwanese\_bankruptcy}\\
\texttt{hr\_analytics} & \texttt{website\_phishing}\\
\bottomrule
\end{tabular}
\end{table}

\paragraph{Full-context references and identical evidence across models.}
For each task, the full-context condition supplies the entire training split as context
(902--100,000 examples), and we use one fixed full-context AUC reference per task and model across all
context sizes. Context selection is model-agnostic: for a given task, context size, and repeat, the
selected examples are constructed once and given to all three TFMs.

\paragraph{Compute and reproducibility.}
The primary TFM evaluations ran on NVIDIA A100 GPUs (40 GiB), one GPU per evaluation. Context
selection is performed offline, before any timing, and is excluded from all reported device
measurements; the on-device profiling methodology and its aggregate results are given in
Appendix~\ref{sec:app-device}.

\section{Complete context-response curves}

Table~\ref{tab:app-budget-summary} reports all primary-context-size distribution summaries. Negative values
are retained because the full context is a reference rather than an oracle. The number within 5 pp is
descriptive only and does not define a universal safety threshold.

\begin{table}[h]
\caption{Context response over the fixed 32-task, three-model support. Each row summarises
96 combinations of tasks and models after averaging over repetitions.}
\label{tab:app-budget-summary}
\centering
\small
\begin{tabular}{rrrrrrr}
\toprule
$C$ & Mean AUC & Mean $\Delta$AUC & Median $\Delta$AUC & SD $\Delta$AUC & $\Delta$AUC range & Within 5 pp\\
\midrule
16  & 0.713 & 14.72 & 14.72 & 8.42 & $-0.82$--32.85 & 15/96\\
32  & 0.742 & 11.83 & 11.65 & 7.63 & $-0.23$--28.90 & 25/96\\
64  & 0.778 &  8.18 &  7.34 & 6.61 & $-3.27$--28.94 & 34/96\\
128 & 0.803 &  5.67 &  4.18 & 5.58 & $-0.57$--25.79 & 58/96\\
256 & 0.824 &  3.66 &  2.40 & 4.47 & $-1.50$--21.54 & 75/96\\
512 & 0.837 &  2.36 &  1.48 & 3.36 & $-0.58$--16.25 & 83/96\\
\bottomrule
\end{tabular}
\end{table}

The matched mean full-reference AUC is 0.860 at every context size. We report the raw observed
profiles rather than imposing a monotone fit; some contain small non-monotone increases (24 of 96
model-specific profiles), which we keep rather than smooth. Figure~\ref{fig:profiles} shows the
three-model-mean grid, while this appendix reports the model-specific summaries. A separate
TabPFN~v2.5 C1024 extension is reported on its smaller support and does not alter the
primary C16--C512 analysis.

\paragraph{Interpreting Hiva's negative $\Delta$AUC.}\label{app:hiva-negative-loss}
The most visible negative cell is \texttt{hiva\_agnostic} at C64. Dataset identity matters here: we
use the three-class TabArena-curated task, not a binary Hiva variant
\citep{guyon2007agnostic,erickson2025tabarena}, which is what makes the class-balance effect below
possible.

Its three-model mean full-context AUC is 0.492, whereas its mean C64 AUC is 0.516, giving a
$\Delta$AUC of $-2.36\pp$. The fixed training split is highly imbalanced, with class counts 28, 2,473, and
62. Because the selector recognises no continuous predictor, its within-class energy-distance score is not
used and the policy falls back to random selection under the approximately equal C64 class
allocation (21, 22, and 21 examples). This rebalancing may partly explain the unusual behaviour:
recent work on prior-data fitted networks finds that severe class imbalance can hurt predictive
performance, and that reducing majority-class examples can improve it \citep{mcdowell2026imbalance}. That prior work studies binary tasks, while Hiva is three-class. We therefore view it as a possible explanation, not a direct one.
The three repeat-level, cross-model mean $\Delta$AUC values are
$-5.67\pp$, $2.36\pp$, and $-3.77\pp$. The negative mean is plausible under balanced selection,
but its large variation across repeats makes it a high-variance result tied to the selection policy,
not evidence that C64 reliably beats full context.

\section{Cross-model agreement and full-context performance control}\label{sec:app-agreement}

Cross-model agreement is evaluated separately for every model pair and context size on the same 32 tasks.
Confidence intervals use 10,000 dataset-level bootstrap resamples; ranks and repeat aggregation are
recomputed within each resample. Across the 18 comparisons, Spearman correlations range from 0.912
to 0.983, and every 95\% interval excludes zero. Table~\ref{tab:app-budget-pairwise} reports all
18 per-context-size estimates and intervals. Holding each model out and comparing it with the
average profile of the other two gives $\rho=0.942$--$0.965$ (leave-one-model-out).

At C128, full-AUC rank agreement is 0.995--0.998. We rank-transform each model's full AUC and $\Delta$AUC,
fit $\Delta$AUC rank on its own full-AUC rank within the model, and correlate the resulting residuals.
Residual agreement remains 0.944--0.975 (Table~\ref{tab:app-headroom}); this is a specific
full-performance adjustment,
not proof that task difficulty is irrelevant.

Repeating this adjustment at every context size leaves the agreement high everywhere: residual
correlations range from 0.909 to 0.982 across the grid, and every 95\% interval stays above zero
(smallest lower bound 0.744), so the effect is not specific to C128. This adjustment only removes the
first-order link with full-context AUC.

\begin{table}[!ht]
\caption{Cross-model agreement at C128 after controlling for each model's full-context performance.
The final column is the Pearson correlation between average-rank OLS residuals.}
\label{tab:app-headroom}
\centering
\small
\begin{tabular}{lccc}
\toprule
Model pair & Full-AUC $\rho$ & Raw-$\Delta$AUC $\rho$ & Residual correlation [95\% CI]\\
\midrule
TabICL~v2--TabPFN~v2.6 & 0.995 & 0.953 & 0.948 [0.870, 0.977]\\
TabICL~v2--TabPFN~v3   & 0.995 & 0.949 & 0.944 [0.871, 0.979]\\
TabPFN~v2.6--TabPFN~v3 & 0.998 & 0.978 & 0.975 [0.916, 0.991]\\
\bottomrule
\end{tabular}
\end{table}

\begin{figure}[!ht]
\centering
\includegraphics[width=\linewidth]{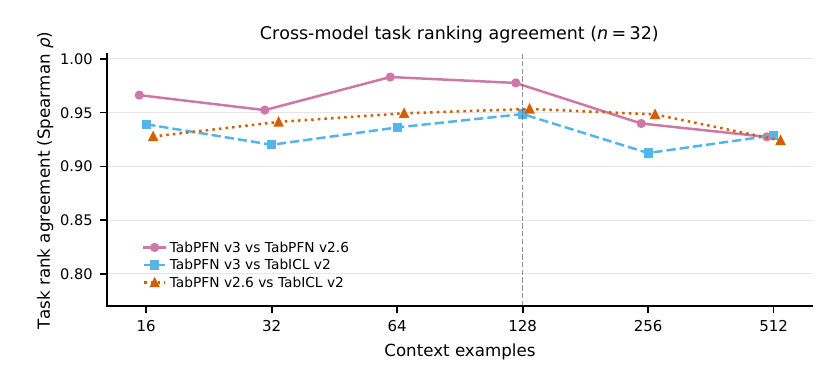}
\caption{Per-context-size cross-model agreement in task ordering on the common 32-task support. At
each context size, every model ranks tasks by $\Delta$AUC and each curve reports the
Spearman correlation for one model pair. The y-axis begins at $\rho=0.77$; $\rho=1$ denotes
identical ordering, not equal AUC or equal $\Delta$AUC magnitude. Table~\ref{tab:app-budget-pairwise}
reports the corresponding 95\% dataset-bootstrap intervals.}
\label{fig:app-budget-agreement}
\end{figure}

At C128, four of the worst five and six of the worst eight tasks recur across all three models.
Appendix Figure~\ref{fig:app-budget-agreement} shows all 18 point estimates for every model pair and context size;
Table~\ref{tab:app-budget-pairwise} reports their intervals, and the remaining tables summarise the
cross-model controls and sensitivities. These analyses characterise agreement among the three
evaluated models, and do not imply that the same ordering would hold for an unseen TFM.

\begin{table}[!ht]
\caption{Per-context-size cross-model task rank agreement on the common 32-task support. Entries are
Spearman $\rho$ [95\% CI] from 10,000 whole-dataset bootstrap resamples. Appendix
Figure~\ref{fig:app-budget-agreement} shows the point estimates without interval bars for
legibility.}
\label{tab:app-budget-pairwise}
\centering
\small
\setlength{\tabcolsep}{4pt}
\begin{tabular}{@{}rccc@{}}
\toprule
$C$ & TabPFN~v3--TabPFN~v2.6 & TabPFN~v3--TabICL~v2 & TabPFN~v2.6--TabICL~v2\\
\midrule
16  & 0.966 [0.891, 0.990] & 0.939 [0.832, 0.978] & 0.928 [0.796, 0.981]\\
32  & 0.952 [0.854, 0.987] & 0.920 [0.790, 0.970] & 0.941 [0.848, 0.975]\\
64  & 0.983 [0.942, 0.994] & 0.936 [0.850, 0.973] & 0.949 [0.880, 0.974]\\
128 & 0.978 [0.920, 0.993] & 0.949 [0.870, 0.982] & 0.953 [0.877, 0.981]\\
256 & 0.940 [0.845, 0.972] & 0.912 [0.753, 0.980] & 0.948 [0.866, 0.978]\\
512 & 0.927 [0.816, 0.975] & 0.929 [0.819, 0.979] & 0.924 [0.813, 0.968]\\
\bottomrule
\end{tabular}
\end{table}

\section{Robustness to context-selection policy}\label{sec:app-selector}

\paragraph{Greedy statistical-energy selection.}
For a selected set $S$ and its class-specific source set $R_k$, we use the empirical energy distance
\begin{align}
\widehat{\mathcal E}(S,R_k)
&=\frac{2}{|S||R_k|}\sum_{x\in S,z\in R_k}\lVert x-z\rVert_2
-\frac{1}{|S|^2}\sum_{x,x'\in S}\lVert x-x'\rVert_2 \nonumber\\
&\quad-\frac{1}{|R_k|^2}\sum_{z,z'\in R_k}\lVert z-z'\rVert_2 .
\label{eq:app-energy}
\end{align}
This statistical distance and its use for representative designs are established
\citep{szekely2013energy,mak2018supportpoints}; our implementation is a discrete greedy heuristic
inspired by that principle, not the Support Points algorithm itself. Within each class, it starts
from the original row closest to the class mean and repeatedly adds the candidate that most reduces
the empirical distance to the class source. Distances use standardised continuous features, while the
selected examples keep every original column. Candidate and reference pools are capped at 300 for
tractability. If a task has no continuous predictor, we use matched random selection under the same
class quota and seed.

We separate two choices in context construction: class allocation and within-class selection. The
four factorial policies use either balanced or class-proportional allocation, combined with either
the energy-distance heuristic or random within-class selection. Quotas are adjusted when a class contains fewer
available examples than requested. Table~\ref{tab:app-selection-factorial} reports the complete C128
comparison over all 32 tasks, three models, and three context-selection repeats.

\begin{table}[h]
\caption{C128 performance under four model-agnostic context-selection policies. Values are mean
$\Delta$AUC in percentage points; lower is better.}
\label{tab:app-selection-factorial}
\centering
\small
\begin{tabular}{llr}
\toprule
Class allocation & Within-class selection & Mean $\Delta$AUC (pp)\\
\midrule
Balanced     & Energy-distance heuristic & 5.667\\
Balanced     & Random        & 6.034\\
Proportional & Energy-distance heuristic & 8.022\\
Proportional & Random        & 8.485\\
\bottomrule
\end{tabular}
\end{table}

Averaging over the two within-class rules, proportional rather than balanced class allocation adds
2.403 pp of $\Delta$AUC; averaging over the two allocation rules, random rather than the
energy-distance heuristic adds 0.415 pp. Within this factorial comparison, class allocation has the
larger descriptive effect. For 12 categorical-only tasks the energy distance is not applied and the
selector falls back to matched random selection, so only the remaining 20 tasks are informative
about energy scoring itself.

To test whether the broad task ordering is created solely by the within-class scoring rule, we also
compare balanced energy-distance and balanced random selection at three context sizes for TabPFN~v2.6.
Table~\ref{tab:app-selector-rank} reports task-level Spearman correlations and paired mean-$\Delta$AUC
differences on the same 32 tasks. The high correlations show that the ordering persists across these
two evaluated policies; they do not imply that selection is irrelevant, and categorical-only tasks
use the matched random fallback under both policies.

\begin{table}[h]
\caption{Multi-context-size comparison of balanced energy-distance and balanced random selection for
TabPFN~v2.6. Positive difference means random selection has higher mean $\Delta$AUC.}
\label{tab:app-selector-rank}
\centering
\small
\begin{tabular}{rcc}
\toprule
Context size & Task-rank $\rho$ [95\% CI] & Random $-$ energy mean $\Delta$AUC (pp)\\
\midrule
64  & 0.978 [0.936, 0.992] & 0.834\\
128 & 0.942 [0.857, 0.977] & 0.430\\
512 & 0.979 [0.934, 0.993] & $-0.010$\\
\bottomrule
\end{tabular}
\end{table}

\section{Classical ML alternative-explanation controls}

We evaluate logistic regression and a shallow decision tree on the same selected examples used by
the TFMs. For each learner and task, we first average $\Delta$AUC over the six context sizes and then
correlate the resulting task rankings; this compares which tasks each learner finds most
context-sensitive on average, not the full shape of their context-response curves. Confidence
intervals resample datasets.

\begin{table}[h]
\caption{Cross-task rank agreement in mean $\Delta$AUC across context sizes. Classical models use the
same selected examples as the TFMs.}
\label{tab:app-classical-control}
\centering
\small
\begin{tabular}{lc}
\toprule
Comparison & Spearman $\rho$ [95\% CI]\\
\midrule
TabPFN~v3--TabPFN~v2.6 & 0.970 [0.896, 0.992]\\
TabPFN~v3--TabICL~v2   & 0.963 [0.876, 0.988]\\
TabPFN~v2.6--TabICL~v2 & 0.957 [0.870, 0.983]\\
Evaluated TFMs--logistic regression & 0.293--0.378 (all intervals include zero)\\
Evaluated TFMs--shallow tree & $-0.058$--$-0.004$ (all intervals include zero)\\
Logistic regression--shallow tree & 0.366 [0.005, 0.651]\\
\bottomrule
\end{tabular}
\end{table}

The evaluated TFMs therefore agree with one another far more than with either classical control:
logistic regression shows only modest correspondence and the shallow tree is near zero. This control
cannot separate a TFM-specific mechanism from generic sample complexity, but it does show that the
shared ordering is not simply a property of the selected rows.

\section{Cross-model profile reuse}
\label{sec:cross-model-profile-reuse}

We summarise each observed response profile at an illustrative tolerance of 5 AUC percentage points using one
of seven categories: C16, C32, C64, C128, C256, C512, or beyond C512. The category is the smallest
observed context size at which $\Delta$AUC is at most 5 points and remains at most 5 points at every larger
observed context size. If no context size meets this rule, the profile is assigned to the beyond-C512
category, which holds 13 task and model combinations.

For every task and ordered pair of models, we use the source model's category as the prediction for
the target model. The prediction is exact in 142 of 192 dataset and ordered model-pair outcomes
(74.0\%), with pair-specific agreement ranging from 68.8\% to 84.4\%. Table~\ref{tab:app-transfer}
compares this result with five profile-free baselines. The modal-category and source-size baselines
are fitted without the evaluated task, while the remaining baselines always predict one fixed
context size. Confidence intervals resample whole datasets, retaining all six model directions together.

\begin{table}[h]
\caption{Same-task cross-model profile reuse at an illustrative 5-point $\Delta$AUC tolerance. Exact
agreement means matching the target model's observed category on the seven-category grid. The final
column reports the paired improvement of profile reuse over each baseline with a 95\% confidence
interval.}
\label{tab:app-transfer}
\centering
\small
\begin{tabular}{lcc}
\toprule
Method & Exact agreement & Reuse improvement [95\% CI]\\
\midrule
Source-model profile & 0.740 & n/a\\
Mode over other tasks & 0.260 & 0.479 [0.302, 0.646]\\
Source-size Ridge & 0.198 & 0.542 [0.375, 0.698]\\
Fixed C128 & 0.250 & 0.490 [0.302, 0.667]\\
Fixed C256 & 0.177 & 0.562 [0.396, 0.719]\\
Fixed C512 & 0.083 & 0.656 [0.510, 0.802]\\
\bottomrule
\end{tabular}
\end{table}

Exact agreement measures whether a method identifies the same observed operating category, not the
utility of a fixed context size. This experiment reuses an already-observed profile for the same task
when switching between the three evaluated TFMs.

\begin{DeviceDraftBlock}
\section{On-device methodology and aggregate measurements}
\label{sec:app-device}

\DeviceDraftMarker
Device measurements use a MacBook Air with an Apple M2, 24\,GB unified memory, eight CPU cores, and
macOS~26.6. TabPFN~v2.6 and TabPFN~v3 execute using the MPS backend; TabICL~v2 executes on CPU after MPS execution failed with an out-of-memory error at model load.
These are the actual execution modes used in this profile. Every model uses eight estimators. For
each dataset, the MacBook profile uses one fixed model-agnostic C128 context, reused across the three
models, while the full-context reference uses the complete fixed training split. Within a task and
model, both conditions use identical preprocessing and an identical stratified query batch of at
most 1{,}000 test rows.

\begin{figure}[!ht]
\DeviceDraftColor
\centering
\includegraphics[width=\linewidth]{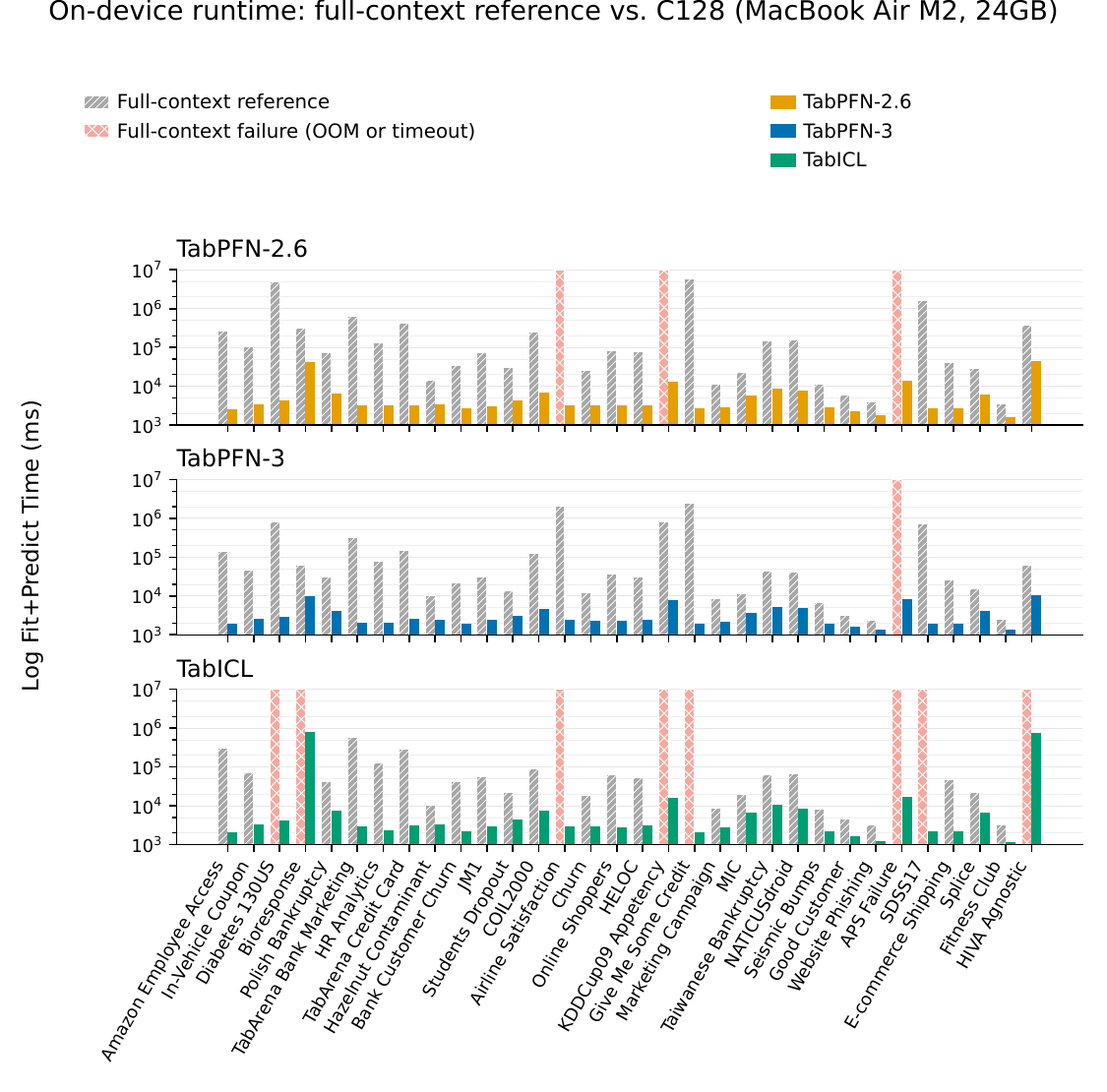}
\caption{Per-dataset MacBook fit+predict time, full-context reference versus C128 context, by
model. Bars are grouped by dataset (ordered by ascending median utility delta, as in the
companion paper's Figure 1; not alphabetical); the hatched grey bar is the full-context
reference and the solid coloured bar is the C128 context. A light-red cross-hatched bar spanning
the full panel height denotes a non-completed full-context run (out-of-memory or the 120-minute
limit) for that task--model pair. All three panels share one log-scaled y-axis.}
\label{fig:app-device-runtime-bars}
\end{figure}

Context construction occurs before device timing. After one untimed warm-up, a combined context
setup and query-prediction pass is timed three times and summarised by its median; every timed repeat
reruns both stages. If the first timed pass exceeds 35 minutes, only that pass is retained. At that
scale the median is set by the run's cost rather than by timing variance, so the two additional
repeats add negligible precision while consuming a large share of the timing budget. Runs are further
capped at 120 minutes per task; a full-context reference that does not finish within this cap is
recorded as non-completing (the cross-hatched bars in Figure~\ref{fig:app-device-runtime-bars}), since
a run this slow is already impractical for that task on this device. Both limits only ever affected
full-context references; C128 runs were never affected. Model loading is excluded, and peak process
RSS is measured over the same execution window for both contexts (Figure~\ref{fig:app-device-memory-bars}).
The paper first forms a C128/full-context ratio within each completed task--model pair and then
summarises those ratios across datasets. Accordingly, P25 and P95 are percentiles across
per-dataset ratios, not percentiles across timing repetitions.

\begin{figure}[!ht]
\DeviceDraftColor
\centering
\includegraphics[width=\linewidth]{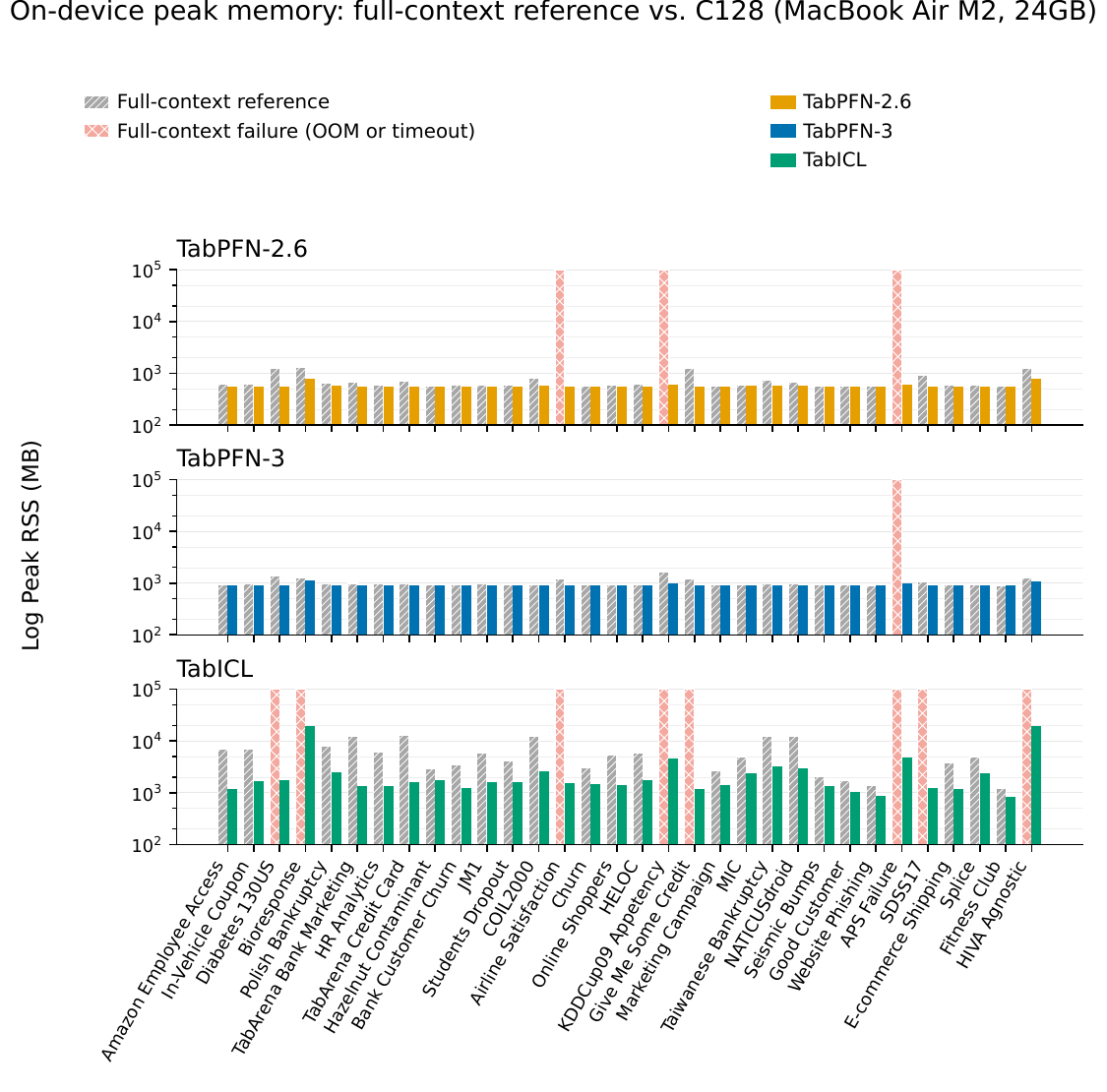}
\caption{Per-dataset MacBook peak process RSS, full-context reference versus C128 context, by
model, in the same layout as Figure~\ref{fig:app-device-runtime-bars}.}
\label{fig:app-device-memory-bars}
\end{figure}

\begin{table}[h]
\DeviceDraftColor
\caption{Aggregate MacBook resource ratios. Time is model-load-excluded combined context setup and
query prediction. A ratio below one favours C128. Geometric means, medians, P25, and P95 are computed
across completed per-dataset ratios.}
\label{tab:app-device-aggregate}
\centering
\small
\setlength{\tabcolsep}{4pt}
\begin{tabular}{lrrrrrrrrrr}
\toprule
& & \multicolumn{4}{c}{Time ratio} & \multicolumn{4}{c}{Peak-RSS ratio}\\
\cmidrule(lr){3-6}\cmidrule(lr){7-10}
Model & $n$ & Geo. & Median & P25 & P95 & Geo. & Median & P25 & P95\\
\midrule
TabPFN v2.6 & 29 & 0.049 & 0.066 & 0.028 & 0.424 & 0.839 & 0.936 & 0.810 & 0.992\\
TabPFN v3   & 31 & 0.060 & 0.089 & 0.022 & 0.521 & 0.910 & 0.950 & 0.911 & 0.992\\
TabICL v2   & 24 & 0.092 & 0.144 & 0.045 & 0.360 & 0.333 & 0.321 & 0.240 & 0.670\\
\bottomrule
\end{tabular}
\end{table}

Table~\ref{tab:app-device-aggregate} reports the aggregate ratios; both C128 ratios are below one in all completed comparisons (29/29, 31/31, and 24/24 by model).
The full-context reference did not complete for 3 TabPFN~v2.6 tasks, 1 TabPFN~v3 task, and 8
TabICL~v2 tasks because of out-of-memory failures or the 120-minute limit; these tasks therefore
have no ratio. The larger TabICL~v2
non-completion count should be read with its execution mode in this profile: TabICL~v2 runs on CPU,
whereas the two TabPFN models run on MPS. Of the 12 non-completing pairs, four were 120-minute
timeouts and the remaining eight were out-of-memory crashes (subprocess exit code 137, except the
TabPFN~v3 \texttt{aps\_failure} run, which recorded no status and is attributed to out-of-memory
because the same task crashed out-of-memory on the other two models). The task datasets that timed out were
\texttt{kddcup09\_appetency} and \texttt{airline\_satisfaction} on TabPFN~v2.6, and
\texttt{airline\_satisfaction} and \texttt{give\_me\_some\_credit} on TabICL~v2. The task datasets
that caused OOM were \texttt{aps\_failure} on TabPFN~v2.6, \texttt{aps\_failure} on
TabPFN~v3, and \texttt{aps\_failure}, \texttt{bioresponse}, \texttt{diabetes\_130us},
\texttt{hiva\_agnostic}, \texttt{kddcup09\_appetency}, and \texttt{sdss17} on TabICL~v2.

\paragraph{Server-side A100 profile.}
We separately measure C128 and full-context inference with one NVIDIA A100 40\,GB GPU per process.
Unlike the MacBook profile, this experiment predicts the complete fixed test split, containing
451--50,000 query rows depending on the dataset. Model loading and offline context construction are
excluded; time combines context setup with prediction over the full query workload. For each task
and model, we first take the median C128/full-context ratio over the three context-selection
repeats, then summarise the 32 task-level ratios with equal task weight. GPU memory is PyTorch peak
allocated memory, rather than the peak process RSS reported for the MacBook profile.
Table~\ref{tab:app-server-aggregate} reports these ratios.

\begin{table}[h]
\DeviceDraftColor
\caption{Server-side C128/full-context resource ratios on an NVIDIA A100 GPU. A ratio below one
favours C128. Combined time includes context setup and prediction over the complete fixed test split
and excludes model loading. Allocated VRAM is PyTorch peak allocated GPU memory.}
\label{tab:app-server-aggregate}
\centering
\small
\setlength{\tabcolsep}{4pt}
\begin{tabular}{llrrrrrr}
\toprule
Model & Metric & $n$ & Geo. mean & Median & P25 & P95 & Ratio $<1$\\
\midrule
TabPFN v2.6 & Combined time  & 32 & 0.268 & 0.297 & 0.178 & 0.930 & 32/32\\
TabPFN v2.6 & Allocated VRAM & 32 & 0.474 & 0.446 & 0.402 & 0.859 & 32/32\\
TabPFN v3   & Combined time  & 32 & 0.499 & 0.657 & 0.363 & 0.975 & 32/32\\
TabPFN v3   & Allocated VRAM & 32 & 0.754 & 0.840 & 0.615 & 1.000 & 28/32\\
TabICL v2   & Combined time  & 32 & 0.492 & 0.591 & 0.434 & 0.854 & 32/32\\
TabICL v2   & Allocated VRAM & 32 & 0.573 & 0.564 & 0.538 & 0.986 & 30/32\\
\bottomrule
\end{tabular}
\end{table}

The MacBook and A100 measurements are separate within-device profiles, not a direct hardware
comparison: their execution backends, software environments, and query workloads differ. In
particular, the MacBook query batch is capped at 1,000 rows, whereas the A100 profile uses the full
test split. The A100 profile completes all full-context comparisons, while the MacBook profile has
12 full-context non-completions; this contrast reinforces that context cost is hardware- and
workload-dependent. We therefore compare C128 with full context only within each profile and do not
compare absolute latency or memory values across the two systems.
\end{DeviceDraftBlock}

\section{Cross-generation robustness}

TabPFN~v2.5 is evaluated separately on the 28 tasks supported by its 50,000-example full-context
limit \citep{grinsztajn2026tabpfn25}. Table~\ref{tab:app-v25} reports $\Delta$AUC relative to
TabPFN~v2.5's own fixed full-context reference.
The four excluded tasks (Airline Satisfaction, APS Failure, Give Me Some Credit, and SDSS17) all
exceed this limit; the exclusions are therefore determined by model support rather than observed
performance.

\begin{table}[!ht]
\caption{TabPFN~v2.5 context response on its matched task support. C1024 is a separate extension with
smaller support and is not combined with the main-grid analysis.}
\label{tab:app-v25}
\centering
\small
\begin{tabular}{rrr}
\toprule
Context size & Tasks & Mean $\Delta$AUC (pp)\\
\midrule
16   & 28 & 14.912\\
32   & 28 & 11.397\\
64   & 28 &  7.910\\
128  & 28 &  5.815\\
256  & 28 &  3.887\\
512  & 28 &  2.319\\
1024 & 13 &  2.102\\
\bottomrule
\end{tabular}
\end{table}

Mean $\Delta$AUC decreases across the main C16--C512 grid, showing the same broad recovery pattern as the
primary models on this separately supported subset. This is a robustness observation, not a causal
comparison of model generations: the versions may differ in architecture, pretraining data, and
inference procedure. The 28-task subset supplements rather than replaces the primary 32-task,
three-model analysis.